%% file: main.tex
\documentclass[letterpaper, 10 pt, journal, twoside]{IEEEtran}
\ifCLASSINFOpdf
  \usepackage[pdftex]{graphicx}
  \graphicspath{{../img/}}
\else
\fi
\usepackage{amsmath}
\usepackage{array}

 \usepackage[caption=true,font=small]{subfig}

\usepackage{stfloats}
\usepackage{url}

\usepackage[monochrome]{color}

\usepackage[noadjust]{cite}  

\makeatletter
\newcommand\footnoteref[1]{\protected@xdef\@thefnmark{\ref{#1}}\@footnotemark}
\makeatother

\usepackage{booktabs} 
\makeatletter
\let\NAT@parse\undefined
\makeatother
\usepackage[hidelinks]{hyperref}
\usepackage{multirow} 
\usepackage[table]{xcolor}  
\usepackage{bm}
\usepackage{multirow} %

\usepackage[capitalize]{cleveref}

\crefname{enumi}{Condition}{Conditions}
\crefname{section}{Sec.}{Secs.}
\Crefname{section}{Section}{Sections}
\Crefname{table}{Table}{Tables}
\crefname{table}{Tab.}{Tabs.}
\crefname{algorithm}{Alg.}{Algs.}

\usepackage{pifont} 
\newcommand{\xmark}{\ding{55}}%
\newcommand{\checkmark}{\ding{51}}%

\graphicspath{{img/}}

\usepackage[nolist]{acronym} 

\input{acronyms.tex}

\def\V{\mathcal{V}}

\def\L{\mathcal{L}}

\def\P{\mathcal{P}}

\def\figvspace{\vspace{-0.3cm}}

\begin{document}

\title{Benchmarking the Sim-to-Real Gap in Cloth Manipulation}

\author{David Blanco-Mulero$^1$, Oriol Barbany$^2$, Gokhan Alcan$^1$, Adrià Colomé$^2$, Carme Torras$^2$, and Ville Kyrki$^1$
\thanks{Manuscript accepted for publication at IEEE Robotics and Automation
Letters.
This research has received funding from: Academy of Finland (grant number 317020), Business Finland (decision 9249/31/2021), 
ERC under the European Union's Horizon 2020 research and innovation programme (grant agreement No. 741930, project CLOTHILDE) 
and European Union's Horizon 2020 research and innovation programme (grant agreement No. 101070600, project SoftEnable).
(\textit{Corresponding author: David Blanco-Mulero}.)
}
\thanks{$^1$~David Blanco-Mulero, Gokhan Alcan and Ville Kyrki are with Department of Electrical Engineering and Automation (EEA), Aalto University, 02150, Espoo, Finland. (e-mail: \texttt{david.blancomulero@aalto.fi})}%
\thanks{$^2$~Oriol Barbany, Adrià Colomé and Carme Torras are with the Institut de Robòtica i Informàtica Industrial, CSIC-UPC, Spain.}%
}


\maketitle
\thispagestyle{empty}
\pagestyle{empty}

\begin{abstract}
Realistic physics engines play a crucial role for learning to manipulate deformable objects such as garments in simulation.
By doing so, researchers can circumvent challenges such as sensing the deformation of the object in the real-world.
In spite of the extensive use of simulations for this task, few works have evaluated the reality gap between deformable object simulators and real-world data.
We present a benchmark \acused{data_set}\acs{data_set} to evaluate the sim-to-real gap in cloth manipulation.
The \acs{data_set} is collected by performing a dynamic as well as a quasi-static cloth manipulation task involving contact with a rigid table.
We use the \acs{data_set} to evaluate the reality gap, computational time, and simulation stability of four popular deformable object simulators: \acs{mujoco}, \acs{bullet}, \acs{flex}, and \acs{sofa}.
Additionally, we discuss the benefits and drawbacks of each simulator.
The benchmark \acs{data_set} is open-source. Supplementary material, videos, and code, can be found at~\texttt{\url{https://sites.google.com/view/cloth-sim2real-benchmark}}.
\end{abstract}



\section{Introduction}
Cloth manipulation is a crucial component in applications ranging from care-giving \cite{clegg_2020_assisted_dressing} and household chores \cite{garcia_2022_cloth_set}, to the textile industry.
Endowing robots with cloth manipulation skills is non-trivial.
First, deformable objects have infinite \acp{dof}, which makes it challenging to represent their state in the world~\cite{arriola_2020_modeling_do_robotic_manipulation_review}.
Second, deformable objects have complex dynamics, which is even further pronounced when performing dynamic manipulation actions
that require acceleration forces to succeed with the task~\cite{ha2021flingbot, Chi_2022_iterative_residual_dynamic_cloth}.
Third, deploying a robot in the real-world presents safety challenges such as damaging the physical system or the environment the robot interacts with.

\begin{figure}
	\centering
	\def\svgwidth{0.9\linewidth}
	{\fontsize{10}{8}
		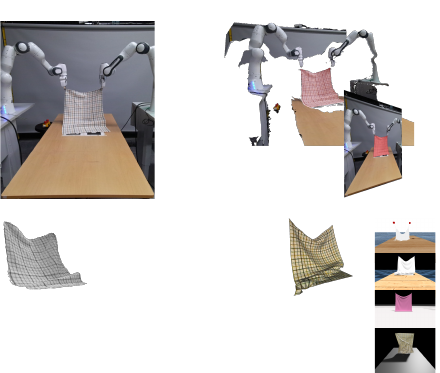}
    \vspace{-0.2cm}
    \caption{
    The real-world cloth manipulation \acs{data_set} was collected, pre-processed and benchmarked against multiple simulation engines, assessing their sim-to-real gap.
    }
	\label{fig:fancy_fig}
    \vspace{-0.5cm}
\end{figure}

Considerable research on cloth manipulation addresses these challenges with the aid of simulation engines~\cite{clegg_2020_assisted_dressing, wu2019learning, yan_2020_contrastive_estimation, seita_bags_2021}.
This relaxes the safety issues and provides a vast amount of trials where the controllers can be evaluated and improved.
However, these simulators approximate the dynamics of the real world, which results in a gap when compared to reality~\cite{collins_2020_benchmarking_sim_manipulation}.
This reality gap becomes even more apparent when performing dynamic cloth manipulation tasks~\cite{jangir2019dynamic, Hietala_2022_ours}. 
Under longer-term prediction the subsequent errors accumulate, widening up the reality gap, which results in a poor sim-to-real transfer.
However, no studies are available that quantify the reality gap when performing dynamic cloth manipulation tasks.

\begin{figure*}[t]
	\centering
	\def\svgwidth{\linewidth}
	{\fontsize{10}{8}
            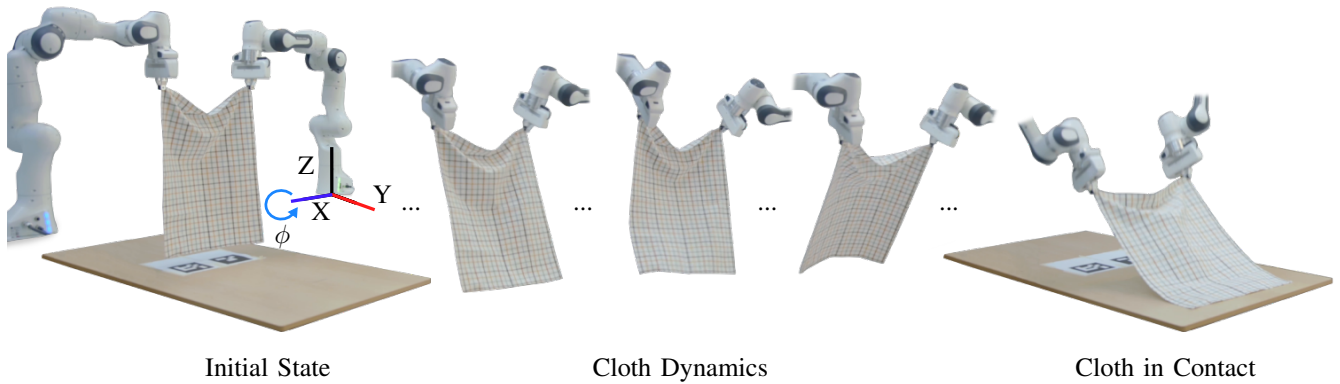
        }
    \vspace{-0.5cm}
    \caption{Robot performing a cloth manipulation task that includes significant features for benchmarking simulation engines. The task exhibits: 
        1) how the high acceleration values of the robot trajectory affect the cloth,
        2) multiple time-steps in which the cloth follows its dynamics,
        3) in-contact with a rigid surface.}
	\label{fig:task_figure}
 \figvspace{}
\end{figure*}

Although the state-of-the-art continues using the available simulators for learning cloth manipulation tasks, the fidelity of simulators for these tasks has not been thoroughly evaluated.
While domain randomisation has been used to obtain more robust controllers that partially alleviate the sim-to-real gap~\cite{blanco2023qdp}, it does not necessarily solve the issue. 
We present a \acs{data_set} for benchmarking cloth manipulation and evaluate the reality gap of current state-of-the-art simulators (\cref{fig:fancy_fig}).
In addition, we provide insights about the available simulators, pointing out their benefits and drawbacks.
Our contributions can be summarised as:
\begin{itemize}
    \item A \acs{data_set} for benchmarking cloth manipulation using cloths from a publicly available benchmarking \acs{data_set}.
    \item Benchmarking the most popular, currently available, physics engines that simulate deformable objects compared to a real-world scenario.
    \item Evaluating the capabilities of physics engines to simulate dynamic in-air manipulation and quasi-static in-contact manipulation of cloths.
\end{itemize}

The work will also enable researchers to evaluate new simulators using the benchmark and the open-source code.

\section{Related Works}\label{sec:related-works}
\subsection{Deformable Object Simulation}
There exists a broad variety of deformable object simulators.
One of their main differences lies on the dynamics model used, ranging from particle-based systems such as the mass-spring (MuJoCo~\cite{todorov_2012_mujoco}) or \ac{pbd} (Flex~\cite{macklin_2014_unified_particle_physics_flex}), to constitutive models such as the \ac{fem} (Bullet~\cite{coumans_2021_bullet}, SOFA~\cite{faure_2012_sofa}).
Although simulators such as Arcsim have been fine-tuned to match the dynamics of fabric materials~\cite{wang_2011_arcsim_cloth}, the reality gap when performing manipulation tasks has not been evaluated.

As a result of the benefits of learning controllers in simulation, recent work has focused on measuring the simulators' accuracy against real-world data. 
Collins \textit{et al.}~\cite{collins_2020_benchmarking_sim_manipulation} benchmarked the accuracy of different simulation engines in a \acused{rigid_object}\acs{rigid_object} manipulation task.
More recently, Acosta \textit{et al.}~\cite{acosta_2022_validating_simulators_impact} measured the error of simulated rigid-body contact after optimising the parameters of different simulators.
However, no prior work has evaluated the reality gap in dynamic deformable object manipulation.

\subsection{Benchmarking Deformable Object Manipulation}
The problem of benchmarking manipulation tasks can be viewed from different perspectives: 1) designing \acp{data_set}~\cite{garcia_2022_cloth_set} and tasks~\cite{garcia_2020_benchmarking_cloth} for benchmarking robotic systems, 2) measuring the performance of multiple algorithms on a task, 3) evaluating the disparity between simulation and real task performance for a given algorithm, and 4) measuring the reality gap between simulation and a real-world \acs{data_set}.

Most works in deformable object manipulation have focused either on 2) evaluating multiple algorithms in a simulation engine~\cite{lin_2021_softgym, chen2023daxbench}, or 3) evaluating the gap when transferring a skill to the real world~\cite{ha2021flingbot, salhotra_2022_learning_dom_reality_gap}.
However, these works do not quantify the reality gap of the simulations used to train the learning algorithms, which can result in poor performance when performing sim-to-real transfer in a zero-shot manner.

More recently, Lim \textit{et al.}~\cite{lim_2022_sim2real2sim_deformable_1d} proposed an approach to learn controllers from real data and simulators fine-tuned with real data for planar cable manipulation, evaluating the reality gap in terms of the cable trajectory.
Similarly, Sundaresan \textit{et al.}~\cite{sundaresan_2022_diffcloud} fine-tuned a differentiable simulator with data from the real-world, evaluating the reality gap in quasi-static tasks.
To the best of our knowledge, our work is the first to study the reality gap in a dynamic cloth manipulation task against a real-world benchmark \acs{data_set}.
In this work we measure the performance of four simulation engines widely used for deformable object manipulation: \acs{mujoco}~\cite{wu2019learning, yan_2020_contrastive_estimation, Hietala_2022_ours}, 
\acs{bullet}~\cite{seita_bags_2021, matas2018sim, antonova_2021_dedo},
\acs{flex}~\cite{ha2021flingbot, blanco2023qdp, lin_2021_softgym}, 
and \acs{sofa}~\cite{jangir2019dynamic, ficuciello_2018_sofa_fem_3d_manipulation}.
Our benchmark is agnostic to the simulator and can be easily applied to forthcoming simulators.

\section{The Benchmark}
The proposed benchmark consists of the following:

\begin{itemize}
    \item a real-world \acs{data_set} composed of \acused{point_cloud}\acp{point_cloud} and RGB-D images at each \acused{timestep}\acs{timestep} of the cloth manipulation, using three cloths with different material properties;
	\item dynamic and quasi-static manipulation tasks performed with a bi-manual system, simulated in four simulation engines;
	\item metrics to evaluate the reality gap of the simulated environment, along with the stability and computational cost of the simulator.
\end{itemize}

\subsection{Task Description}

We propose a fabric placement manipulation task performed by a bi-manual system that involves two pre-defined trajectories (see~\cref{fig:task_figure}).
The first trajectory consists of a dynamic motion of the fabric.
The second trajectory brings the garment in contact with a rigid surface and then drags it through the planar surface.
The objective of the two trajectories is to evaluate two different dynamics: 1) the dynamics of the fabric without contact, and 2) the dynamics of the contact between the garment and a rigid object.

The goal of the fabric placement task is to end in a flattened configuration starting from a position free of contact.
In order to focus on the accuracy of the simulation, the task assumes a successful grasp state. Thus, the fabric starts in a grasped position, where two corners of a rectangular piece of cloth are grasped by a manipulator using a pinch grasp \cite{borras_2020_grasp_cloth_analysis}.
We decide to use a pinch grasp as this does not require any additional \acused{set_up}\acs{set_up} such as those required for interfacing and simulating, e.g., touch-based sensors~\cite{proesmans_2023_tactile_manipulation_cloth_unfolding}.

In order to place the cloth in a flattened configuration, dynamic motions can control the fabric outside of the working space of the manipulators while
efficiently placing the cloth flat in a single attempt.
Thus, we design a fling motion \cite{ha2021flingbot} and define it with a quintic polynomial
\begin{equation}\label{eq:quintic-poly}
x(t) = a_0 + a_1t + a_2t^2+a_3t^3+a_4t^4+a_5t^5,
\end{equation}
which is detailed in~\cref{sec:real-world-data_set}.
When performing a highly dynamic motion the fabric suffers an abrupt deformation.
This is challenging to simulate due to the inertia forces generated by high accelerations and high number of \acp{dof} of the garment.
Therefore, it is a great candidate trajectory for evaluating the reality gap. In addition, to evaluate the capability of different physics engines to simulate frictional and inertia forces, we design a quasi-static motion which consists in entering in contact with the rigid surface by slowly lowering and dragging the fabric.

\begin{figure}
\vspace{0.15cm}  
     \centering
     \subfloat[Dynamic motion trajectory with $n_Y=4$, $n_Z=3$ and $n_\phi=3$.]{\def\svgwidth{0.96\linewidth}
	{\fontsize{8}{8}
		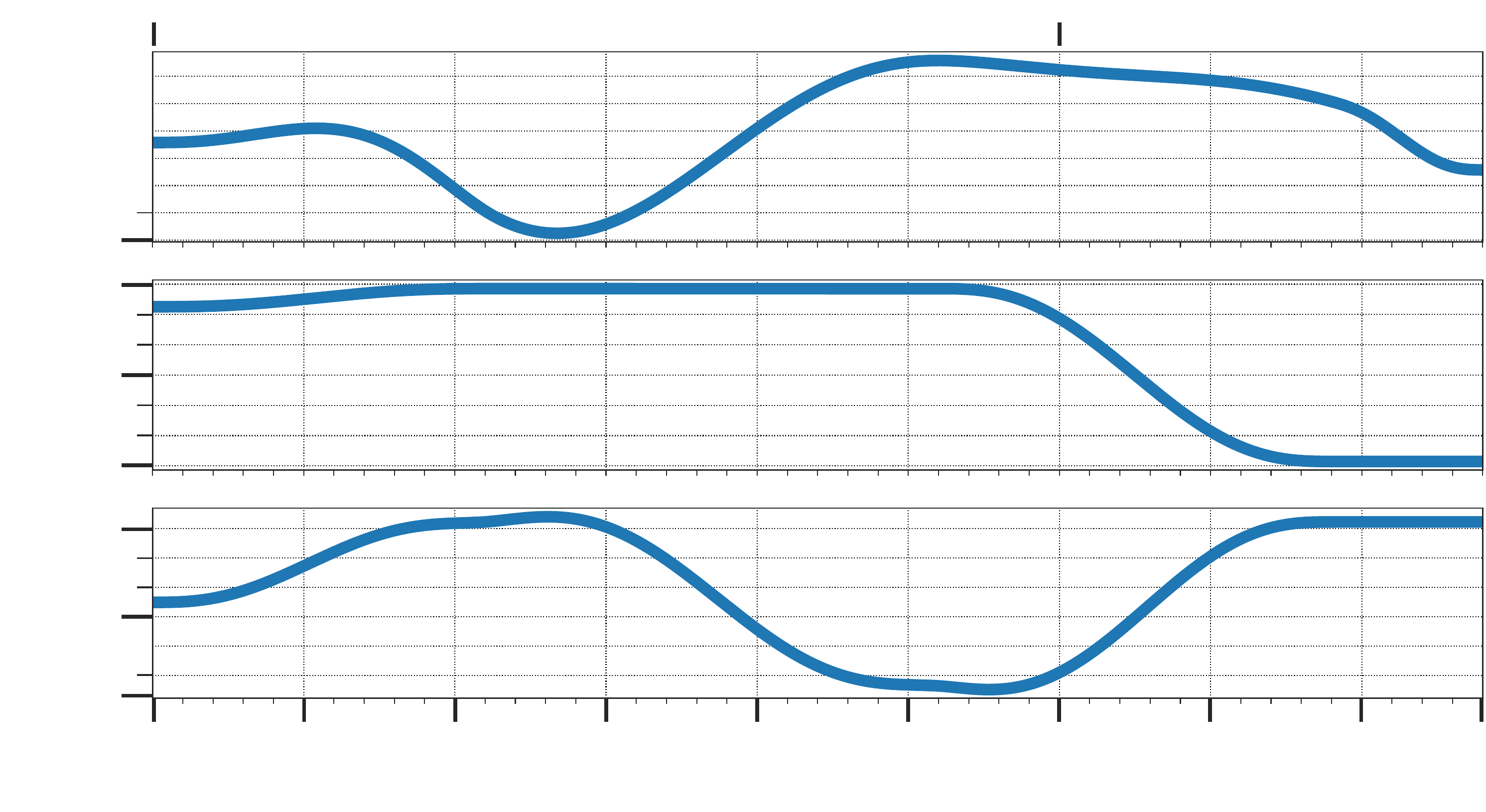}
  \label{fig:trajectory_fig_dynamic}}\\
  \subfloat[Quasi-static motion trajectory with $n_Y=2$ and $n_Z=2$.]
  {\def\svgwidth{0.96\linewidth}
	{\fontsize{8}{8}
		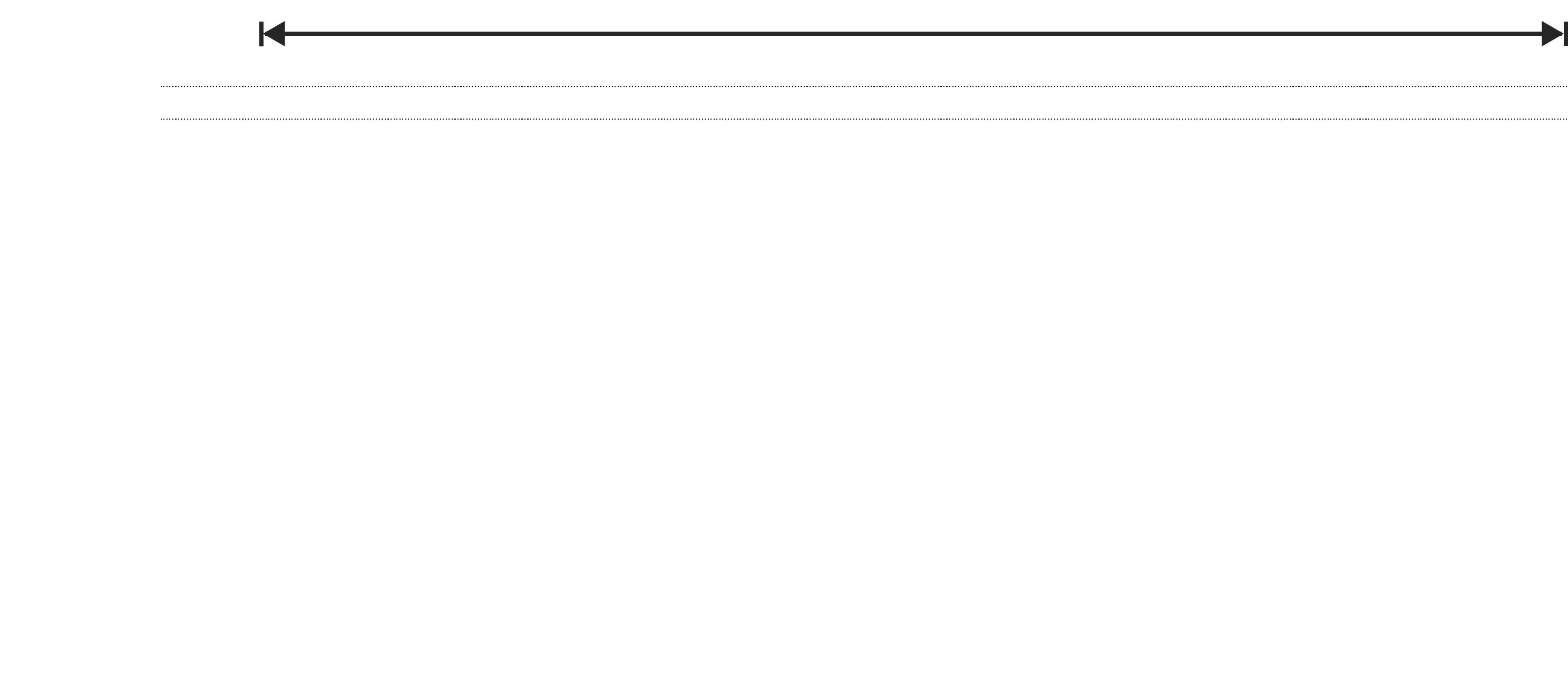}
  \label{fig:trajectory_fig_quasi_static}}
    \caption{Quintic trajectories composed of different number of via-points for the Y--axis, Z--axis, and roll angle.}
    \label{fig:trajectory_fig}
    \figvspace{}
\end{figure}

\subsection{Real-World Dataset}\label{sec:real-world-data_set}
Our \acs{data_set} is collected using three different cloths from the public household \acs{data_set} \cite{garcia_2022_cloth_set}: a towel rag, a linen rag, and a chequered rag. 
The garments have a size of 50 $\times$ 70 cm, each with different weight and elasticity, providing a variety that helps assess the ability of the engines to simulate different fabric materials and dynamics.
We decided to use these cloths from the \acs{data_set} as they can be easily lifted by two robotic manipulators and placed on a flat surface.
In addition, the fabrics have a rectangular shape rather than square, which results in larger deformations of the non-manipulated corners of the cloth when a high-velocity is applied.
As shown in~\cref{sec:experiments}, the towel and the chequered rags have a similar final configuration after the fling motion. However, the linen rag, which is more brittle, is partially folded. By contrast, all fabrics exhibit a similar final configuration after the quasi-static motion.

We use two Franka Emika Panda robots to perform the quintic trajectories.
The dynamic trajectory performs a motion on the $YZ$ axes and the roll angle $\phi$, keeping the other axes fixed throughout the trajectory (see~\cref{fig:trajectory_fig_dynamic}).
By contrast, the quasi-static trajectory performs a motion only on the $YZ$ axes (see~\cref{fig:trajectory_fig_quasi_static}).
Each trajectory is computed using multiple via-points, where the number of via-points is $n_Y = 4$, $n_Z = 3$ and $n_\phi=3$ for each the dynamic trajectory, and $n_Y = 2$, $n_Z = 2$ for the quasi-static trajectory.
For each via-point we define a quintic polynomial, where the coefficients of the polynomial are computed following~\cite{spong2020robot}.
The starting and final velocity and acceleration values, as well as the time of the trajectory for each via-point, are defined empirically.
The position and velocity trajectories for each axis are shown in~\cref{fig:trajectory_fig}.
During the trajectory, both robots have the $X$--axis fixed at $51$ cm from their origin.
Since one of the manipulators is rotated $180$ degrees with respect to the other, its roll angle is inverted.

In the dynamic motion (\cref{fig:trajectory_fig_dynamic}), we distinguish between the phase where the cloth undergoes its natural dynamics and when it makes contact with the surface. The cloth dynamics phase, concluding approximately 3 seconds after starting the trajectory, is used in the benchmarking. Although the trajectory remains consistent across all trials and materials, the cloth contacts the table at slightly different time steps due to inherent randomness in cloth behaviour and material differences\footnote{These values can be found in our open-source code.}. Consequently, we manually refine the change of phase \acs{timestep} for each case.
By contrast, in the quasi-static trajectory we evaluate both the time instant where the fabric enters into contact as well as the entire contact phase.

The \acp{point_cloud} of the \acs{data_set} are captured using a Microsoft Azure Kinect RGB-D sensor.
The RGB-D images have a dimensionality of $1280 \times 720$, and are captured at a frame rate of 30 fps.
To compare how well the garments resemble reality in a simulator, we propose to compare the dense \acs{point_cloud} $\P$ obtained by the sensor in the real setup with the meshes $\V$ of the garment provided by the simulator.
This enables to quantitatively compare the reality gap, as we can measure the distance between the simulated and real fabric points, rather than performing a qualitative comparison by e.g. comparing their deformation using RGB images.

To obtain the \acp{point_cloud} $\P$ we use the real-world RGB-D images, as well as the position of the camera w.r.t. the manipulators and the intrinsic and extrinsic camera parameters.
First, we segment the RGB images with MiVOS~\cite{cheng2021mivos}, which allows to interactively refine the segmentation on individual frames and obtain temporally coherent results. This enables filtering out points that are not part of the garment.
Since the positions of the robots and the boundary positions of the garment are known, we use these positions to filter the points.
Then, we discard points further away from their neighbours compared to the average. This is performed by applying an statistical outlier removal.
Finally, we need to account for the possibly different coordinate systems used across simulators. To achieve this, we define the appropriate coordinate transformation matrices and apply them to convert the simulated meshes into the observation space.

\subsection{Simulation Engine Set-up}
To benchmark the sim-to-real gap in the cloth manipulation tasks we design a framework that is agnostic to the simulation engine and share it open-source\footnote{\scriptsize{\url{https://sites.google.com/view/cloth-sim2real-benchmark}}}.
In order to benchmark a simulation engine using our \acs{data_set}, the simulator needs to have the following capabilities:
\begin{itemize}
    \item simulation of both rigid and deformable objects,
    \item control over the cloth points,
    \item information about the position of the mesh points of the garment,
    \item adjustable frequency of the simulation engine.
\end{itemize}
The simulation scene is comprised of the same elements as the real-world \acs{data_set}: a \acs{rigid_object} surface, the fabric to manipulate, and two manipulators.
Given the limitations present in some simulators (see~\cref{sec:simulation-engines}), we consider that a robot is not available in the simulator and assume that only a pinch grasp is available using a dummy manipulator.

The simulated manipulators must take as input the desired target Cartesian coordinate position.
The trajectories are given in Cartesian coordinates and calculated according to the specific simulator $\Delta t$. Thus, the trajectories are agnostic to the simulator frequency.
To accurately follow the trajectory of the real-world \acs{data_set}, the simulator must not modify the \acs{data_set} trajectories or repeat the same action in case that frame-skips are used, as often done in \acs{mujoco} or \acs{bullet}.

Given the variety of dynamic models used to approximate the behaviour of cloths in simulation engines, there is no restriction on the cloth parameters.
In addition, our benchmark can be used to fine-tune simulator parameters such as the damping coefficient or stiffness that better approximate the dynamics of the garments.

\subsection{Performance Metrics}
The objectives of our metrics are to: 1) qualitatively measure the reality gap, 2) evaluate the stability of the simulated cloth, and 3) assess the capability of using the simulated scenes in real-time control (hardware-in-the-loop).

There are multiple candidate metrics for measuring the distance between two \acp{point_cloud}, such as the \ac{cd}, the \ac{hd}, or the Earth-mover distance.
We select the \acs{cd} and \acs{hd} as they do not require point correspondences between the real \acs{point_cloud} and the simulated mesh, are efficient, and permutation invariant.
We use the unidirectional (also known as one-way or one-sided) \acs{cd} and \acs{hd} to address different mesh resolutions, and incomplete \acp{point_cloud} due to self-occlusions, as done in previous works facing the same issues on clothes~\cite{sundaresan_2022_diffcloud,huang_2023_cloth_reconstruction_tracking}.
For a \acs{point_cloud} $\P_t$ and a simulated mesh with vertices $\V_t$, the \ac{cd} used for evaluating the reality gap is defined as
\begin{align}
    \text{CD}(\V_t, \P_t):=\frac{1}{|\V_t|}\sum_{v \in \V_t} \min_{p\in \P_t}\|v-p\|_1 \,.
\end{align}
The unidirectional \ac{hd} with $\ell_1$ norm is defined as
\begin{align}
    \text{HD}(\V_t, \P_t):= \max_{v \in \V_t} \min_{p\in \P_t}\|v-p\|_1\,.
\end{align}
The \ac{hd} is closely related to the \ac{cd} and greater by definition, as it corresponds to the largest error, whereas the \ac{cd} is an average of errors.
Both metrics typically use the squared Euclidean distance. However, we empirically find that the error values obtained with the Manhattan distance are more representative. The reason for that is that the $\ell_1$ norm is more robust to outliers, an observation consistent with the use of the un-squared $\ell_2$ norm as an evaluation measure in previous works \cite{occupancy_networks,DVR}.
Note that $|\V_t|\ll |\P_t|$, further motivating the use of the $\ell_1$ norm, which could cause the metric to blow up in the presence of a few extreme values.

To evaluate the modelling of cloth dynamics, we use the recorded trajectory before the collision, as detailed in~\cref{sec:real-world-data_set}. For this purpose, we report in Table~\ref{tab:benchmark-cloud-comparison} the average of the Chamfer and Hausdorff distances between the simulated mesh vertices and \acp{point_cloud} up to the change of phase \acs{timestep}, denoted $\text{CD}_d$ and $\text{HD}_d$.

The quasi-static trajectory is used in its entirety to evaluate the simulation of contacts in the absence of fast dynamic motions. The reported metrics in this case are $\text{CD}_q$ and $\text{HD}_q$, representing the average of distances across all \acp{timestep}.

The \ac{hd} is closely related to the \ac{cd}, and, by definition, it has a value greater than or equal to it.
Both distances determine point correspondences by finding the closest pairs between sets. 
However, the \ac{cd} reports the average of distances and hence has higher tolerance for outliers, while the \ac{hd} is a stricter metric that focuses on the maximum dissimilarity.
Overall, both metrics offer complementary and valuable information about the reality gap.
One of the drawbacks of both the \ac{cd} and \ac{hd} is that they do not consider the connectivity of the mesh \cite{atlasnet}. 
However, in our case, the mesh connectivity is already enforced by the physics simulator.

We provide as a reference the error metrics between each of the target \acp{point_cloud} in~\cref{tab:benchmark-cloud-comparison}.
The table measures the difference in their deformation and serves as a guide to understand the metric values in~\cref{sec:results-sim-to-real-gap}.

To evaluate the simulator stability, we apply a moving average filter to the simulated vertices and compute the difference between the filtered and non-filtered vertices as
\begin{equation}
    \L_\text{s} = \dfrac{1}{N} \sum^{N-1}_{t=1} \left| \dfrac{ \V_{t-1} + \V_t + \V_{t+1}}{3} - \V_t \right|.
\end{equation}

Finally, to measure the capability of using the simulators in real-time control, we measure the computational time taken to perform a single simulation step and contrast it against the simulator frequency and error metrics aforementioned.

\begin{table*}[ht]
\vspace{0.2cm}  
\caption{Comparison of the evaluated simulators: \acs{mujoco}, \acs{bullet}, \acs{flex} and \acs{sofa}.
Here, we compare if the simulator: 1) has visual feedback (RGB-D), 2) has robotic systems, 3) the type of grasp, 4) the numerical integrator, and 5) CPU or GPU acceleration.
Specifically for deformable objects, we contrast whether 1) meshes can be used (variable);  and 2) the dynamics model used for deformable objects.
The type of grasp is considered as points (P) or lines (L) \cite{borras_2020_grasp_cloth_analysis}.
}
\label{tab:physics-comparison}
\begin{center}
\resizebox{%
      \ifdim\width>\linewidth
        \linewidth
      \else
        \width
      \fi
    }{!}{
\begin{tabular}{lccccccc}
\toprule
 & \multicolumn{4}{c}{\textbf{Simulation Generic}} & \multicolumn{2}{c}{\textbf{Deformable Objects}} \\
 Physics Simulator & RGB-D & Robot Integration & Grasp &  CPU/GPU &  Shape  & Dynamics Model  \\
\midrule
\acs{mujoco} \cite{todorov_2012_mujoco} & \checkmark & \checkmark  & P/L & CPU \& GPU & Variable\footnoteref{mujoco3_footnote}  & Mass-Spring \\
\acs{bullet} \cite{coumans_2021_bullet} & \checkmark & \checkmark   & P/L & CPU & Variable  &  \acs{pbd} / \acs{fem}  \\
\acs{flex} \cite{macklin_2014_unified_particle_physics_flex, lin_2021_softgym} & \checkmark & \checkmark\footnoteref{flex_footnote} & P/L &  GPU & Variable & \acs{pbd}  \\
\acs{sofa} \cite{faure_2012_sofa} & RGB & \xmark  & P & CPU \& GPU    & Variable & Mass-Spring / \acs{fem} \\
\bottomrule
\end{tabular}
}
\end{center}
\end{table*}

\input{tabs/benchmark_cloud_comparison}

\begin{table*}
\vspace{0.2cm}  
\caption{Quantitative result showing the mean and standard deviation for the \acf{cd} and \acf{hd} for three rags: towel, chequered and linen; over three real world \acp{data_set} of dynamic and quasi-static tasks for each fabric, and 20 different random seeds in the physic engines \acs{mujoco}, \acs{bullet}, \acs{flex} and \acs{sofa}.}
\label{tab:quantitative-results}
\begin{center}
\begin{small}
\begin{sc}
\resizebox{%
      \ifdim\width>\linewidth
        \linewidth
      \else
        \width
      \fi
    }{!}{
\begin{tabular}{l c c c c c}
\toprule
Rag & Metric & \acs{mujoco} & \acs{bullet} & \acs{flex} & \acs{sofa}\\
\toprule
\multirow{4}{*}{\textbf{Towel}} & $\text{CD}_d$ &  \underline{0.079} $\pm$ \underline{0.031}  &  0.155 $\pm$ 0.093  &  0.168 $\pm$ 0.129  &  \textbf{0.078 $\pm$ 0.029}  \\ 
 & $\text{HD}_d$ &  \textbf{0.167 $\pm$ 0.037}  &  0.206 $\pm$ 0.083  &  0.277 $\pm$ 0.158  &  \underline{0.194} $\pm$ \underline{0.074}  \\ 
 \cmidrule{2-6} 
 & $\text{CD}_q$ &  \textbf{0.079 $\pm$ 0.027}  &  0.097 $\pm$ 0.045  &  \underline{0.080} $\pm$ \underline{0.021}  &  0.089 $\pm$ 0.022  \\ 
 & $\text{HD}_q$ &  0.182 $\pm$ 0.040  &  0.243 $\pm$ 0.078  &  \textbf{0.169 $\pm$ 0.024}  &  \underline{0.174} $\pm$ \underline{0.032}  \\ 
 \midrule 
\multirow{4}{*}{\textbf{Cheq.}} & $\text{CD}_d$ &  \textbf{0.067 $\pm$ 0.026}  &  0.119 $\pm$ 0.060  &  0.164 $\pm$ 0.134  &  \underline{0.068} $\pm$ \underline{0.024}  \\ 
 & $\text{HD}_d$ &  \textbf{0.154 $\pm$ 0.035}  &  0.242 $\pm$ 0.063  &  0.280 $\pm$ 0.180  &  \underline{0.178} $\pm$ \underline{0.051}  \\ 
 \cmidrule{2-6} 
 & $\text{CD}_q$ &  \underline{0.076} $\pm$ \underline{0.025}  &  0.094 $\pm$ 0.034  &  \textbf{0.072 $\pm$ 0.019}  &  0.102 $\pm$ 0.034  \\ 
 & $\text{HD}_q$ &  \underline{0.186} $\pm$ \underline{0.055}  &  0.243 $\pm$ 0.070  &  \textbf{0.171 $\pm$ 0.024}  &  0.198 $\pm$ 0.020  \\ 
 \midrule 
\multirow{4}{*}{\textbf{Linen}} & $\text{CD}_d$ &  \underline{0.071} $\pm$ \underline{0.031}  &  0.116 $\pm$ 0.054  &  0.160 $\pm$ 0.131  &  \textbf{0.061 $\pm$ 0.024}  \\ 
 & $\text{HD}_d$ &  \underline{0.154} $\pm$ \underline{0.033}  &  0.235 $\pm$ 0.076  &  0.281 $\pm$ 0.186  &  \textbf{0.150 $\pm$ 0.064}  \\ 
 \cmidrule{2-6} 
 & $\text{CD}_q$ &  0.083 $\pm$ 0.037  &  0.078 $\pm$ 0.023  &  \underline{0.075} $\pm$ \underline{0.023}  &  \textbf{0.073 $\pm$ 0.021}  \\ 
 & $\text{HD}_q$ &  0.177 $\pm$ 0.068  &  0.182 $\pm$ 0.024  &  \underline{ 0.148} $\pm$ \underline{0.045}  &  \textbf{0.137 $\pm$ 0.038}  \\
 
 \bottomrule
\end{tabular}
}
\end{sc}
\end{small}
\end{center}
\vspace{-0.2cm}
\end{table*}

\section{Experiments}\label{sec:experiments}

\subsection{Simulation Engines}\label{sec:simulation-engines}
The simulation engines\footnote{The experiments were performed using \acs{mujoco} v3.1.1, \acs{bullet} v3.26, \acs{flex} v1.2 and \acs{sofa} v23.06.} selected for our experiments and their main differences are depicted in~\cref{tab:physics-comparison}. 
\subsubsection{Visual Feedback}
Starting from a more generic point of view, we note that all simulators provide visual feedback, such as an RGB-D camera.
However, setting up specific camera properties, such as the intrinsics o{\color{red}r} extrinsics of a camera, is not straightforward for neither \acs{flex} nor \acs{sofa}, which require modifying the source code of the simulation engine.
Similarly, there are available solutions for domain randomisation in \acs{mujoco} and \acs{bullet} \cite{robosuite2020}, while \acs{flex} and \acs{sofa} do require additional software such as Blender to randomise properties such as texture or colour of objects.

\subsubsection{Robot Integration and Type of Grasp}
It is important to note that \acs{sofa} does not simulate robot systems and these are not available by default in \acs{flex}\footnote{\label{flex_footnote}IsaacSim which incorporates \acs{flex} does simulate robotic systems.}.
The lack of an end-effector will result in a larger impact on the sim-to-real gap when learning visual feedback controllers.
Regarding the type of grasping, the simulation engines that have robotic models, and therefore grippers, enable both point (P) and line (L) grasps~\cite{borras_2020_grasp_cloth_analysis}. Moreover, although \acs{sofa} lacks robotic models, the grasping technique could be modified to perform line grasps.

\subsubsection{GPU acceleration}
In terms of GPU acceleration, \acs{bullet} is a CPU-based simulation engine, while \acs{flex} supports only CUDA simulation. By contrast, both \acs{mujoco} and \acs{sofa} support both CPU and GPU-based simulation, although in our benchmark we only use the CPU-based for a fair evaluation against the other CPU-based simulators.


\subsubsection{Deformable Object Shape}
Regarding the shape of deformable objects, all simulators provide the capability of loading 3D meshes\footnote{\label{mujoco3_footnote}The latest version of \acs{mujoco} can load 3D meshes.}.
Although \acs{flex} is limited by default to rectangular shapes, defining garments by their width and length, recent work by Ha \textit{et~al.}~\cite{ha2021flingbot} has extended \acs{flex} to non-rectangular shapes.

\subsubsection{Deformable Object Dynamics Model}
As discussed in~\cref{sec:related-works}, the dynamics model used for simulating a cloth approximates its behaviour and has an effect on the reality gap.
\acs{mujoco} models cloths as mass-spring systems, which are connected by joints. The simulator allows for the definition of shear and stretch joints, enabling more complex behaviours.
\acs{bullet} uses PBD by default to model object dynamics. However, this must be switched to FEM for simulating deformable objects.
Similarly, \acs{sofa} provides an FEM implementation to simulate object deformation.
Finally, \acs{flex} uses PBD to model the object dynamics.

\subsection{Evaluating the Sim-to-Real Gap}\label{sec:results-sim-to-real-gap}
Prior to evaluating the reality gap for each physics engine, we optimise the simulator cloth parameters that best fit the behaviour of each \acs{data_set} using the standard optimisation procedures of \ac{bo}~\cite{garnett2023bayesian} and the \ac{cmaes}~\cite{hansen2011cma}.
For doing so, we run 500 sweeps of both \ac{bo} and \ac{cmaes} in each simulation engine, where we minimise the CD against each fabric and quintic trajectory, and keep the random seed constant. Once the number of sweeps is reached, or the optimisation converges, we select the parameters leading to the lowest distance over \ac{bo} or \ac{cmaes}.
The specific parameters used for each simulator can be found in our open-source code.
We use the default numerical integrators for \acs{bullet} and \acs{flex}, semi-implicit Euler for \acs{mujoco}, and implicit conjugate gradient for \acs{sofa}.

We evaluate the reality gap against each fabric using 20 random seeds per simulation engine.
The quantitative results for each fabric, task and simulator are reported in~\cref{tab:quantitative-results}.
Overall, all engines perform similarly for the quasi-static task, where all distances are in the same order of magnitude for both \acs{cd} and \acs{hd}.
In contrast, the difference in performance in the dynamic task is more noticeable across engines.
Both \acs{mujoco} and \acs{sofa} present distances  two times lower than those in \acs{bullet} and \acs{flex}.

In general, the values for all metrics are comparable, or greater, than the distances between the chequered and linen rags shown in~\cref{tab:benchmark-cloud-comparison}.

\begin{figure*}[t]
\vspace{0.25cm}  
	\centering
	\def\svgwidth{0.85\linewidth}
	{\fontsize{8}{7}
		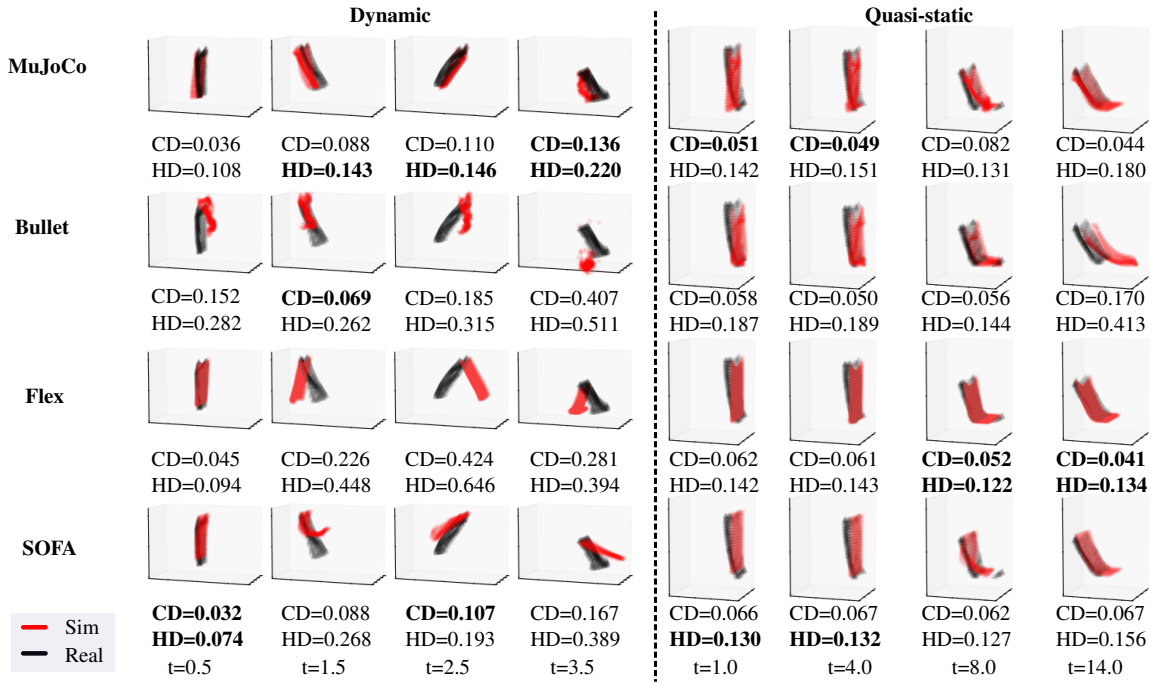}
        \caption{Qualitative and quantitative results of the simulated cloth mesh (red) in the selected simulators and one of the towel \acp{point_cloud} from the \acs{data_set} (black) at different \acp{timestep} of the dynamic (left) and quasi-static (right) trajectories. The \acf{cd} and \acf{hd} are shown below the figure at each \acs{timestep} of the simulator.
        }
	\label{fig:qualitative_sim_real_comparison}
\vspace{-0.4cm}
\end{figure*}

In addition, we qualitatively assess the reality gap in each simulator
by visualising both simulated and \acs{data_set} cloth \acp{point_cloud} in~\cref{fig:qualitative_sim_real_comparison}, where we randomly select one of the simulations obtained for the optimal parameters of the towel rag.
The figure also shows the distances associated to each \acs{timestep}, which helps to further understand the metrics.
We can notice that \acs{mujoco} is the only engine that closely follows the dynamic trajectory. In contrast, although \acs{bullet} presents low values at $t=1.5$, the identified parameters do not produce a stable simulation, resulting in an inconsistent result at $t=3.5$.
On the other hand, all simulators are able to closely match the quasi-static trajectory, where \acs{flex} has the lowest error at the final \acs{timestep}.

\subsection{Simulator Stability, Computational Time and Reality Gap}
\begin{figure}
\vspace{0.1cm}
    \centering
	\def\svgwidth{0.95\linewidth}
	{\fontsize{9}{9}
  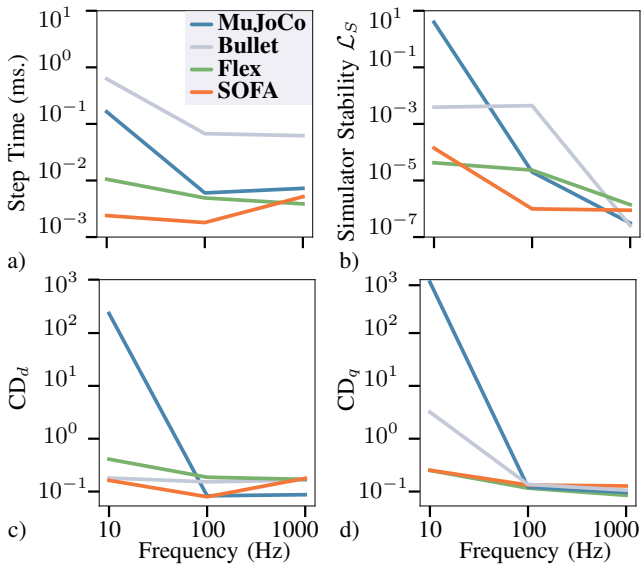}
        \caption{Comparison of a) step time taken per simulation step in seconds, b) simulator stability ($L_\text{s}$), c) dynamic task $\text{CD}_d$, and d) quasi-static task $\text{CD}_q$; at the frequencies of 10, 100 and 1000 Hz for each of the simulation engines evaluated using the towel from the benchmark.}
    \label{fig:exp-freq-chamfer-time}
    \figvspace{} 
\end{figure}

We report the relationship between reality gap, computational time\footnote{All experiments were run using an Intel i7-10875H and an RTX 2070.} and simulator frequency in~\cref{fig:exp-freq-chamfer-time}, where we report the average \acs{cd} for the dynamic and quasi-static tasks.
We only report the stability values for the dynamic task due to the difficulty of the engines to simulate this trajectory.
We used the benchmark data from the towel rag with 10 random seeds per simulation engine, while keeping the same fine-tuned simulation parameters as in~\cref{sec:results-sim-to-real-gap}. We selected $10$, $100$ and $1000$ Hz as frequencies for each engine.

As shown by~\cref{fig:exp-freq-chamfer-time} b) both \acs{bullet} and \acs{mujoco} become unstable when using a low frequency, while \acs{flex} and \acs{sofa} are more consistent at different frequencies.
We can notice a drastic improvement in performance for \acs{mujoco} when increasing the frequency in both~\cref{fig:exp-freq-chamfer-time} c) and~\cref{fig:exp-freq-chamfer-time} d).
By contrast, \acs{flex} and \acs{sofa} present similar values at different frequencies.
Although higher frequencies result in a more stable computation of the system dynamics, there is no improvement in the distance, or even some detrimental performance. This suggests that the physics engines are quite sensitive to the cloth parameters. Therefore, all engines need to be fine-tuned for the specific simulation frequency.

The computational time taken per simulation step is depicted in~\cref{fig:exp-freq-chamfer-time} a).
We can notice that, for the case of \acs{bullet} and \acs{mujoco}, if the simulator is unstable the time drastically increases.
Given that the time taken per simulation step for 100Hz for all simulators is in the order of milliseconds, it is unfeasible to perform real-time dynamic manipulation with hardware-in-the-loop.
Similarly, although the simulation step time for 10Hz in \acs{flex} and \acs{sofa} is lower, and they are more stable than \acs{mujoco} and \acs{bullet}, their \acs{cd} is still pretty high for hardware-in-the-loop manipulation.

\section{Discussion and Future Work}
Our results show that the largest reality gap results from performing dynamic cloth manipulation.
Although the impact of this gap is less pronounced for tasks that do not require high accelerations, for instance, diagonal folding~\cite{matas2018sim}, techniques such as sim-to-real-to-sim \cite{lim_2022_sim2real2sim_deformable_1d} might be beneficial for closing this gap \cite{hofer_2021_sim2real_applications_and_challenges}.

Although \acs{sofa} presents lower errors on both tasks, its lack of robotic models does not make it an effective simulation engine for learning robotic controllers that require visual feedback.
Regarding \acs{bullet}, the identification of the system parameters requires greater efforts to produce reasonable results as the simulation parameters also affect the grasping of the fabrics, which is why in our benchmark it performed poorly for the dynamic task.
On the other hand, although \acs{flex} was able to produce a swing motion, it was not able to match the real cloth behaviour. In addition, it is restricted to GPU acceleration.
Given the lower distances in both dynamic and quasi-static manipulation tasks shown by \acs{mujoco}, as well as its capability of integrating robotic models, and availability of both CPU and GPU acceleration, we recommend \acs{mujoco} for learning cloth manipulation tasks in a simulation engine.

Although our \acs{data_set} is designed with three cloths with different properties, none of these fabrics had a strong resistance to deformation.
During our research we found out that only \acs{bullet} and \acs{mujoco} were able to approximate the behaviour of stiff garments.
The evaluation of the reality gap for stiff cloths and other types of garments such as jeans and t-shirts remains as future work. 

\section{Conclusion}
In this letter, we presented a benchmark that evaluates the reality gap of physics engines simulating cloth manipulation tasks and evaluated four well-established open-source simulation engines: \acs{mujoco}, \acs{bullet}, \acs{flex}, and \acs{sofa}.
Our benchmark \acs{data_set} was collected using three cloths from a public household \acs{data_set}, each with different material properties, in both a dynamic and quasi-static manipulation task.
The benchmark \acs{data_set} provides the \acp{point_cloud} of the post-processed cloths, as well as the trajectory performed by the robots.
Our experiments evaluate qualitatively and quantitatively the discrepancy between the benchmark \acs{data_set} for each fabric, task, and simulated cloth.
Furthermore, we analysed the computational time taken for each simulator at different frequencies, along with their stability and the reality gap.
Our results show that all engines are able to produce low errors for the quasi-static task.
However, although none of the simulators was able to precisely match the dynamic manipulation task, \acs{mujoco} performed the best at closely following the dynamic trajectory.
The remaining reality gap emphasises that, in order to transfer controllers learnt in simulation to the real world, techniques such as domain randomisation, real-to-sim or real-time visual feedback are required.

Our benchmark was designed to aid researchers in cloth manipulation by depicting the current capabilities of simulation engines.
Our work also provides the open-source code, which enables evaluating the reality gap of other simulation engines, as well as performing other tasks and trajectories using the same \acs{set_up} as the one depicted in this letter.
We foresee that the next generation of simulators will have a lower reality gap evaluated against these benchmarks, leading to controllers that match more faithfully the behaviour learnt in simulation when applied in the real world.


\section*{Acknowledgement}
The authors would like to acknowledge the computational resources provided by the Aalto Science-IT project.

\bibliography{benchmarking, manipulation, simulators, computer_vision}
\bibliographystyle{IEEEtran}
\end{document}

%% file: acronyms.tex
\newacro{dof}[DoF]{Degree of Freedom}
\acrodefplural{dof}{Degrees of Freedom}
\newacro{mujoco}[MuJoCo]{MuJoCo}
\newacro{bullet}[Bullet]{Bullet}
\newacro{sofa}[SOFA]{SOFA}
\newacro{flex}[Flex]{Flex}
\newacro{isaacsim}[IsaacSim]{IsaacSim}
\newacro{physx}[PhysX]{PhysX}
\newacro{dcm}[DCM]{Dynamic Cloth Manipulation}
\newacro{fem}[FEM]{Finite Element Method}
\newacro{pbd}[PBD]{Position Based Dynamics}
\newacro{cd}[CD]{Chamfer Distance}
\newacro{hd}[HD]{Hausdorff Distance}
\newacro{point_cloud}[point cloud]{point cloud}
\newacro{timestep}[time step]{time step}
\newacro{timepoint}[time point]{time point}
\newacro{data_set}[dataset]{dataset}
\newacro{rigid_object}[rigid-object]{rigid-object}
\newacro{set_up}[set-up]{set-up}
\newacro{cmaes}[CMA-ES]{Covariance Matrix Adaptation Evolution Strategy}
\newacro{bo}[BO]{Bayesian Optimisation}

%% file: img/latest_version.pdf_tex
\begingroup%
  \makeatletter%
  \providecommand\color[2][]{%
    \errmessage{(Inkscape) Color is used for the text in Inkscape, but the package 'color.sty' is not loaded}%
    \renewcommand\color[2][]{}%
  }%
  \providecommand\transparent[1]{%
    \errmessage{(Inkscape) Transparency is used (non-zero) for the text in Inkscape, but the package 'transparent.sty' is not loaded}%
    \renewcommand\transparent[1]{}%
  }%
  \providecommand\rotatebox[2]{#2}%
  \newcommand*\fsize{\dimexpr\f@size pt\relax}%
  \newcommand*\lineheight[1]{\fontsize{\fsize}{#1\fsize}\selectfont}%
  \ifx\svgwidth\undefined%
    \setlength{\unitlength}{210.05865863bp}%
    \ifx\svgscale\undefined%
      \relax%
    \else%
      \setlength{\unitlength}{\unitlength * \real{\svgscale}}%
    \fi%
  \else%
    \setlength{\unitlength}{\svgwidth}%
  \fi%
  \global\let\svgwidth\undefined%
  \global\let\svgscale\undefined%
  \makeatother%
  \begin{picture}(1,0.84916678)%
    \lineheight{1}%
    \setlength\tabcolsep{0pt}%
    \put(0,0){\includegraphics[width=\unitlength,page=1]{latest_version.pdf}}%
    \put(-0.00418407,0.81999769){\color[rgb]{0,0,0}\makebox(0,0)[lt]{\lineheight{1.25}\smash{\begin{tabular}[t]{l}1. Record Dataset\end{tabular}}}}%
    \put(0.50083772,0.82090089){\color[rgb]{0,0,0}\makebox(0,0)[lt]{\lineheight{1.25}\smash{\begin{tabular}[t]{l}2. Pre-Process Data-set\end{tabular}}}}%
    \put(-0.00258023,0.35825765){\color[rgb]{0,0,0}\makebox(0,0)[lt]{\lineheight{1.25}\smash{\begin{tabular}[t]{l}3. Measure Reality Gap in Simulated Environments\end{tabular}}}}%
    \put(0.25853247,0.23052531){\color[rgb]{0,0,0}\makebox(0,0)[lt]{\lineheight{1.25}\smash{\begin{tabular}[t]{l}Compute Distance\end{tabular}}}}%
    \put(0.20061584,0.13127733){\color[rgb]{0,0,0}\makebox(0,0)[lt]{\lineheight{1.25}\smash{\begin{tabular}[t]{l}Real\end{tabular}}}}%
    \put(0.58646226,0.13127733){\color[rgb]{0,0,0}\makebox(0,0)[lt]{\lineheight{1.25}\smash{\begin{tabular}[t]{l}Sim\end{tabular}}}}%
    \put(0,0){\includegraphics[width=\unitlength,page=2]{latest_version.pdf}}%
  \end{picture}%
\endgroup%

%% file: img/fig2_v2.pdf_tex
\begingroup%
  \makeatletter%
  \providecommand\color[2][]{%
    \errmessage{(Inkscape) Color is used for the text in Inkscape, but the package 'color.sty' is not loaded}%
    \renewcommand\color[2][]{}%
  }%
  \providecommand\transparent[1]{%
    \errmessage{(Inkscape) Transparency is used (non-zero) for the text in Inkscape, but the package 'transparent.sty' is not loaded}%
    \renewcommand\transparent[1]{}%
  }%
  \providecommand\rotatebox[2]{#2}%
  \newcommand*\fsize{\dimexpr\f@size pt\relax}%
  \newcommand*\lineheight[1]{\fontsize{\fsize}{#1\fsize}\selectfont}%
  \ifx\svgwidth\undefined%
    \setlength{\unitlength}{1539.17522761bp}%
    \ifx\svgscale\undefined%
      \relax%
    \else%
      \setlength{\unitlength}{\unitlength * \real{\svgscale}}%
    \fi%
  \else%
    \setlength{\unitlength}{\svgwidth}%
  \fi%
  \global\let\svgwidth\undefined%
  \global\let\svgscale\undefined%
  \makeatother%
  \begin{picture}(1,0.29349102)%
    \lineheight{1}%
    \setlength\tabcolsep{0pt}%
    \put(0,0){\includegraphics[width=\unitlength,page=1]{fig2_v2.pdf}}%
    \put(0.15856221,0.00847){\color[rgb]{0,0,0}\makebox(0,0)[lt]{\lineheight{1.25}\smash{\begin{tabular}[t]{l}$\text{Initial State}$\end{tabular}}}}%
    \put(0.2267631,0.15668606){\color[rgb]{0,0,0}\makebox(0,0)[lt]{\lineheight{1.25}\smash{\begin{tabular}[t]{l}Z\end{tabular}}}}%
    \put(0.28141421,0.13550725){\color[rgb]{1,0.10196078,0.10196078}\makebox(0,0)[lt]{\lineheight{1.25}\smash{\begin{tabular}[t]{l}Y\end{tabular}}}}%
    \put(0.23640689,0.1229437){\color[rgb]{0.25882353,0.10196078,1}\makebox(0,0)[lt]{\lineheight{1.25}\smash{\begin{tabular}[t]{l}X\end{tabular}}}}%
    \put(0.20935678,0.10622211){\color[rgb]{0.10980392,0.52941176,1}\makebox(0,0)[lt]{\lineheight{1.25}\smash{\begin{tabular}[t]{l}$\phi$\end{tabular}}}}%
    \put(0.44263177,0.00879103){\color[rgb]{0,0,0}\makebox(0,0)[lt]{\lineheight{1.25}\smash{\begin{tabular}[t]{l}$\text{Cloth Dynamics}$\end{tabular}}}}%
    \put(0.79661811,0.00847){\color[rgb]{0,0,0}\makebox(0,0)[lt]{\lineheight{1.25}\smash{\begin{tabular}[t]{l}$\text{Cloth in Contact}$\end{tabular}}}}%
    \put(0.30220628,0.1303793){\color[rgb]{0,0,0}\makebox(0,0)[lt]{\lineheight{1.25}\smash{\begin{tabular}[t]{l}$\text{...}$\end{tabular}}}}%
    \put(0.42824928,0.1303793){\color[rgb]{0,0,0}\makebox(0,0)[lt]{\lineheight{1.25}\smash{\begin{tabular}[t]{l}$\text{...}$\end{tabular}}}}%
    \put(0.56343639,0.1303793){\color[rgb]{0,0,0}\makebox(0,0)[lt]{\lineheight{1.25}\smash{\begin{tabular}[t]{l}$\text{...}$\end{tabular}}}}%
    \put(0.6965289,0.1303793){\color[rgb]{0,0,0}\makebox(0,0)[lt]{\lineheight{1.25}\smash{\begin{tabular}[t]{l}$\text{...}$\end{tabular}}}}%
    \put(0,0){\includegraphics[width=\unitlength,page=2]{fig2_v2.pdf}}%
  \end{picture}%
\endgroup%

%% file: img/pos_quintic_traj.pdf_tex
\begingroup%
  \makeatletter%
  \providecommand\color[2][]{%
    \errmessage{(Inkscape) Color is used for the text in Inkscape, but the package 'color.sty' is not loaded}%
    \renewcommand\color[2][]{}%
  }%
  \providecommand\transparent[1]{%
    \errmessage{(Inkscape) Transparency is used (non-zero) for the text in Inkscape, but the package 'transparent.sty' is not loaded}%
    \renewcommand\transparent[1]{}%
  }%
  \providecommand\rotatebox[2]{#2}%
  \newcommand*\fsize{\dimexpr\f@size pt\relax}%
  \newcommand*\lineheight[1]{\fontsize{\fsize}{#1\fsize}\selectfont}%
  \ifx\svgwidth\undefined%
    \setlength{\unitlength}{1431.54244995bp}%
    \ifx\svgscale\undefined%
      \relax%
    \else%
      \setlength{\unitlength}{\unitlength * \real{\svgscale}}%
    \fi%
  \else%
    \setlength{\unitlength}{\svgwidth}%
  \fi%
  \global\let\svgwidth\undefined%
  \global\let\svgscale\undefined%
  \makeatother%
  \begin{picture}(1,0.54634807)%
    \lineheight{1}%
    \setlength\tabcolsep{0pt}%
    \put(0,0){\includegraphics[width=\unitlength,page=1]{pos_quintic_traj.pdf}}%
    \put(0.49763793,0.00184187){\color[rgb]{0,0,0}\makebox(0,0)[lt]{\lineheight{1.25}\smash{\begin{tabular}[t]{l}Time (seconds)\end{tabular}}}}%
    \put(0.1830942,0.03376565){\color[rgb]{0,0,0}\makebox(0,0)[lt]{\lineheight{1.25}\smash{\begin{tabular}[t]{l}0.5\end{tabular}}}}%
    \put(0.09136856,0.03376565){\color[rgb]{0,0,0}\makebox(0,0)[lt]{\lineheight{1.25}\smash{\begin{tabular}[t]{l}0.0\end{tabular}}}}%
    \put(0.28453468,0.03376565){\color[rgb]{0,0,0}\makebox(0,0)[lt]{\lineheight{1.25}\smash{\begin{tabular}[t]{l}1.0\end{tabular}}}}%
    \put(0.38656188,0.03385774){\color[rgb]{0,0,0}\makebox(0,0)[lt]{\lineheight{1.25}\smash{\begin{tabular}[t]{l}1.5\end{tabular}}}}%
    \put(0.4880025,0.03376565){\color[rgb]{0,0,0}\makebox(0,0)[lt]{\lineheight{1.25}\smash{\begin{tabular}[t]{l}2.0\end{tabular}}}}%
    \put(0.69147023,0.03376565){\color[rgb]{0,0,0}\makebox(0,0)[lt]{\lineheight{1.25}\smash{\begin{tabular}[t]{l}3.0\end{tabular}}}}%
    \put(0.79349744,0.03376565){\color[rgb]{0,0,0}\makebox(0,0)[lt]{\lineheight{1.25}\smash{\begin{tabular}[t]{l}3.5\end{tabular}}}}%
    \put(0.89493793,0.03376565){\color[rgb]{0,0,0}\makebox(0,0)[lt]{\lineheight{1.25}\smash{\begin{tabular}[t]{l}4.0\end{tabular}}}}%
    \put(0.96538098,0.03385774){\color[rgb]{0,0,0}\makebox(0,0)[lt]{\lineheight{1.25}\smash{\begin{tabular}[t]{l}4.5\end{tabular}}}}%
    \put(0.59002974,0.03376565){\color[rgb]{0,0,0}\makebox(0,0)[lt]{\lineheight{1.25}\smash{\begin{tabular}[t]{l}2.5\end{tabular}}}}%
    \put(-0.00137117,0.07562025){\color[rgb]{0,0,0}\rotatebox{90}{\makebox(0,0)[lt]{\lineheight{1.25}\smash{\begin{tabular}[t]{l}$\phi$ (rad)\end{tabular}}}}}%
    \put(-0.0006276,0.24694593){\color[rgb]{0,0,0}\rotatebox{90}{\makebox(0,0)[lt]{\lineheight{1.25}\smash{\begin{tabular}[t]{l}$Z$ (m)\end{tabular}}}}}%
    \put(0.00002729,0.39533519){\color[rgb]{0,0,0}\rotatebox{90}{\makebox(0,0)[lt]{\lineheight{1.25}\smash{\begin{tabular}[t]{l}$Y$ (m)\end{tabular}}}}}%
    \put(0.02870838,0.49285166){\color[rgb]{0,0,0}\makebox(0,0)[lt]{\lineheight{1.25}\smash{\begin{tabular}[t]{l}0.2\end{tabular}}}}%
    \put(0.01947582,0.43477096){\color[rgb]{0,0,0}\makebox(0,0)[lt]{\lineheight{1.25}\smash{\begin{tabular}[t]{l}-0.1\end{tabular}}}}%
    \put(0.01886186,0.38114817){\color[rgb]{0,0,0}\makebox(0,0)[lt]{\lineheight{1.25}\smash{\begin{tabular}[t]{l}-0.4\end{tabular}}}}%
    \put(0.01908015,0.18501376){\color[rgb]{0,0,0}\makebox(0,0)[lt]{\lineheight{1.25}\smash{\begin{tabular}[t]{l}-2.5\end{tabular}}}}%
    \put(0.0104097,0.12620511){\color[rgb]{0,0,0}\makebox(0,0)[lt]{\lineheight{1.25}\smash{\begin{tabular}[t]{l}-3.25\end{tabular}}}}%
    \put(0.01900511,0.07129185){\color[rgb]{0,0,0}\makebox(0,0)[lt]{\lineheight{1.25}\smash{\begin{tabular}[t]{l}-4.0\end{tabular}}}}%
    \put(0,0){\includegraphics[width=\unitlength,page=2]{pos_quintic_traj.pdf}}%
    \put(0.0279307,0.34554213){\color[rgb]{0,0,0}\makebox(0,0)[lt]{\lineheight{1.25}\smash{\begin{tabular}[t]{l}1.0\end{tabular}}}}%
    \put(0.02825132,0.28905461){\color[rgb]{0,0,0}\makebox(0,0)[lt]{\lineheight{1.25}\smash{\begin{tabular}[t]{l}0.7\end{tabular}}}}%
    \put(0.02816605,0.22810472){\color[rgb]{0,0,0}\makebox(0,0)[lt]{\lineheight{1.25}\smash{\begin{tabular}[t]{l}0.4\end{tabular}}}}%
    \put(0,0){\includegraphics[width=\unitlength,page=3]{pos_quintic_traj.pdf}}%
    \put(0.3737154,0.53442364){\color[rgb]{0.14901961,0.14901961,0.14901961}\makebox(0,0)[lt]{\lineheight{1.25}\smash{\begin{tabular}[t]{l}Cloth Dynamics\end{tabular}}}}%
    \put(0.77813954,0.53319231){\color[rgb]{0.14901961,0.14901961,0.14901961}\makebox(0,0)[lt]{\lineheight{1.25}\smash{\begin{tabular}[t]{l}In-Contact\end{tabular}}}}%
  \end{picture}%
\endgroup%

%% file: img/pos_quasi_static_traj.pdf_tex
\begingroup%
  \makeatletter%
  \providecommand\color[2][]{%
    \errmessage{(Inkscape) Color is used for the text in Inkscape, but the package 'color.sty' is not loaded}%
    \renewcommand\color[2][]{}%
  }%
  \providecommand\transparent[1]{%
    \errmessage{(Inkscape) Transparency is used (non-zero) for the text in Inkscape, but the package 'transparent.sty' is not loaded}%
    \renewcommand\transparent[1]{}%
  }%
  \providecommand\rotatebox[2]{#2}%
  \newcommand*\fsize{\dimexpr\f@size pt\relax}%
  \newcommand*\lineheight[1]{\fontsize{\fsize}{#1\fsize}\selectfont}%
  \ifx\svgwidth\undefined%
    \setlength{\unitlength}{1427.99258423bp}%
    \ifx\svgscale\undefined%
      \relax%
    \else%
      \setlength{\unitlength}{\unitlength * \real{\svgscale}}%
    \fi%
  \else%
    \setlength{\unitlength}{\svgwidth}%
  \fi%
  \global\let\svgwidth\undefined%
  \global\let\svgscale\undefined%
  \makeatother%
  \begin{picture}(1,0.43041042)%
    \lineheight{1}%
    \setlength\tabcolsep{0pt}%
    \put(0,0){\includegraphics[width=\unitlength,page=1]{pos_quasi_static_traj.pdf}}%
    \put(0.54572753,0.41870254){\color[rgb]{0.14901961,0.14901961,0.14901961}\makebox(0,0)[lt]{\lineheight{1.25}\smash{\begin{tabular}[t]{l}In-Contact\end{tabular}}}}%
    \put(0,0){\includegraphics[width=\unitlength,page=2]{pos_quasi_static_traj.pdf}}%
    \put(0.49746626,0.00184645){\color[rgb]{0,0,0}\makebox(0,0)[lt]{\lineheight{1.25}\smash{\begin{tabular}[t]{l}Time (seconds)\end{tabular}}}}%
    \put(0.09018689,0.03384956){\color[rgb]{0,0,0}\makebox(0,0)[lt]{\lineheight{1.25}\smash{\begin{tabular}[t]{l}0.0\end{tabular}}}}%
    \put(0.219106,0.03384956){\color[rgb]{0,0,0}\makebox(0,0)[lt]{\lineheight{1.25}\smash{\begin{tabular}[t]{l}2.0\end{tabular}}}}%
    \put(0.34720846,0.03394188){\color[rgb]{0,0,0}\makebox(0,0)[lt]{\lineheight{1.25}\smash{\begin{tabular}[t]{l}4.0\end{tabular}}}}%
    \put(0.47555375,0.03384956){\color[rgb]{0,0,0}\makebox(0,0)[lt]{\lineheight{1.25}\smash{\begin{tabular}[t]{l}6.0\end{tabular}}}}%
    \put(0.72724189,0.03384956){\color[rgb]{0,0,0}\makebox(0,0)[lt]{\lineheight{1.25}\smash{\begin{tabular}[t]{l}10.0\end{tabular}}}}%
    \put(0.85545383,0.03384956){\color[rgb]{0,0,0}\makebox(0,0)[lt]{\lineheight{1.25}\smash{\begin{tabular}[t]{l}12.0\end{tabular}}}}%
    \put(0.95810344,0.03394188){\color[rgb]{0,0,0}\makebox(0,0)[lt]{\lineheight{1.25}\smash{\begin{tabular}[t]{l}14.0\end{tabular}}}}%
    \put(0.60377944,0.03384956){\color[rgb]{0,0,0}\makebox(0,0)[lt]{\lineheight{1.25}\smash{\begin{tabular}[t]{l}8.0\end{tabular}}}}%
    \put(0,0){\includegraphics[width=\unitlength,page=3]{pos_quasi_static_traj.pdf}}%
    \put(0.00093827,0.11272316){\color[rgb]{0,0,0}\rotatebox{90}{\makebox(0,0)[lt]{\lineheight{1.25}\smash{\begin{tabular}[t]{l}$Z$ (m)\end{tabular}}}}}%
    \put(-0.00050607,0.26148135){\color[rgb]{0,0,0}\rotatebox{90}{\makebox(0,0)[lt]{\lineheight{1.25}\smash{\begin{tabular}[t]{l}$Y$ (m)\end{tabular}}}}}%
    \put(0.01887117,0.37008237){\color[rgb]{0,0,0}\makebox(0,0)[lt]{\lineheight{1.25}\smash{\begin{tabular}[t]{l}-0.6\end{tabular}}}}%
    \put(0.01899081,0.30101513){\color[rgb]{0,0,0}\makebox(0,0)[lt]{\lineheight{1.25}\smash{\begin{tabular}[t]{l}-0.3\end{tabular}}}}%
    \put(0.02601008,0.247259){\color[rgb]{0,0,0}\makebox(0,0)[lt]{\lineheight{1.25}\smash{\begin{tabular}[t]{l} 0.0\end{tabular}}}}%
    \put(0.02746673,0.19370723){\color[rgb]{0,0,0}\makebox(0,0)[lt]{\lineheight{1.25}\smash{\begin{tabular}[t]{l}1.0\end{tabular}}}}%
    \put(0.02778812,0.13707926){\color[rgb]{0,0,0}\makebox(0,0)[lt]{\lineheight{1.25}\smash{\begin{tabular}[t]{l}0.7\end{tabular}}}}%
    \put(0.02770266,0.07597782){\color[rgb]{0,0,0}\makebox(0,0)[lt]{\lineheight{1.25}\smash{\begin{tabular}[t]{l}0.4\end{tabular}}}}%
    \put(0,0){\includegraphics[width=\unitlength,page=4]{pos_quasi_static_traj.pdf}}%
  \end{picture}%
\endgroup%

%% file: tabs/benchmark_cloud_comparison.tex
\begin{table}
\vspace{0.1cm}
\caption{Error metrics between the \acs{data_set} \acp{point_cloud}. The table rows refer to the source \acs{point_cloud} $\V$ and the columns to the target \acs{point_cloud} $\P$ for both the Chamfer Distance (CD) and Hausdorff Distance (HD).}
\label{tab:benchmark-cloud-comparison}
\begin{center}
\begin{small}
\begin{sc}
\resizebox{\linewidth}{!}{
\begin{tabular}{l c c c c}
\toprule
Rag   & Metric &  \textbf{Towel} &  \textbf{Cheq.} & \textbf{Linen} \\
\toprule
\multirow{4}{*}{\textbf{Towel}} & $\text{CD}_d$ & - &0.023$\pm$0.000& 0.050$\pm$0.001\\
 & $\text{HD}_d$ &-&0.119$\pm$0.000& 0.163$\pm$0.008\\
 \cmidrule{2-5}
& $\text{CD}_q$ &-& 0.022$\pm$0.000&0.018$\pm$0.000\\
  & $\text{HD}_q$ &-&0.136$\pm$0.003&0.161$\pm$0.003\\
\midrule
\multirow{4}{*}{\textbf{Cheq.}}  & $\text{CD}_d$ & 0.026$\pm$0.000&-&0.036$\pm$0.000\\
 & $\text{HD}_d$ & 0.087$\pm$0.001 & - &0.124$\pm$0.004\\
  \cmidrule{2-5}
 & $\text{CD}_q$ &0.024$\pm$0.000&-&0.023$\pm$0.000\\
 & $\text{HD}_q$ & 0.068$\pm$0.001&-&0.088$\pm$0.002\\
\midrule
\multirow{4}{*}{\textbf{Linen}} & $\text{CD}_d$ &0.036$\pm$0.001&0.054$\pm$0.001 &- \\
 & $\text{HD}_d$ & 0.133$\pm$0.002&0.145$\pm$0.005 &-\\
  \cmidrule{2-5}
& $\text{CD}_q$ & 0.022$\pm$0.000&0.022$\pm$0.000&- \\
 & $\text{HD}_q$ & 0.121$\pm$0.001&0.125$\pm$0.000&-\\
\bottomrule
\end{tabular}
}
\end{sc}
\end{small}
\end{center}
\vspace{-0.2cm}
\end{table}

%% file: img/timesteps_sim_real.pdf_tex
\begingroup%
  \makeatletter%
  \providecommand\color[2][]{%
    \errmessage{(Inkscape) Color is used for the text in Inkscape, but the package 'color.sty' is not loaded}%
    \renewcommand\color[2][]{}%
  }%
  \providecommand\transparent[1]{%
    \errmessage{(Inkscape) Transparency is used (non-zero) for the text in Inkscape, but the package 'transparent.sty' is not loaded}%
    \renewcommand\transparent[1]{}%
  }%
  \providecommand\rotatebox[2]{#2}%
  \newcommand*\fsize{\dimexpr\f@size pt\relax}%
  \newcommand*\lineheight[1]{\fontsize{\fsize}{#1\fsize}\selectfont}%
  \ifx\svgwidth\undefined%
    \setlength{\unitlength}{1076.69043353bp}%
    \ifx\svgscale\undefined%
      \relax%
    \else%
      \setlength{\unitlength}{\unitlength * \real{\svgscale}}%
    \fi%
  \else%
    \setlength{\unitlength}{\svgwidth}%
  \fi%
  \global\let\svgwidth\undefined%
  \global\let\svgscale\undefined%
  \makeatother%
  \begin{picture}(1,0.58105071)%
    \lineheight{1}%
    \setlength\tabcolsep{0pt}%
    \put(0,0){\includegraphics[width=\unitlength,page=1]{timesteps_sim_real.pdf}}%
    \put(0.01241868,0.10925086){\color[rgb]{0,0,0}\makebox(0,0)[lt]{\lineheight{1.25}\smash{\begin{tabular}[t]{l}\textbf{\acs{sofa}}\end{tabular}}}}%
    \put(0.01485307,0.24065107){\color[rgb]{0,0,0}\makebox(0,0)[lt]{\lineheight{1.25}\smash{\begin{tabular}[t]{l}\textbf{\acs{flex}}\end{tabular}}}}%
    \put(0.00589914,0.386063){\color[rgb]{0,0,0}\makebox(0,0)[lt]{\lineheight{1.25}\smash{\begin{tabular}[t]{l}\textbf{\acs{bullet}}\end{tabular}}}}%
    \put(0,0.52530555){\color[rgb]{0,0,0}\makebox(0,0)[lt]{\lineheight{1.25}\smash{\begin{tabular}[t]{l}\textbf{\acs{mujoco}}\end{tabular}}}}%
    \put(0.29464931,0.56976033){\color[rgb]{0,0,0}\makebox(0,0)[lt]{\lineheight{1.25}\smash{\begin{tabular}[t]{l}\textbf{Dynamic}\end{tabular}}}}%
    \put(0.73911838,0.56924153){\color[rgb]{0,0,0}\makebox(0,0)[lt]{\lineheight{1.25}\smash{\begin{tabular}[t]{l}\textbf{Quasi-static}\end{tabular}}}}%
    \put(0,0){\includegraphics[width=\unitlength,page=2]{timesteps_sim_real.pdf}}%
    \put(0.04901554,0.04086527){\color[rgb]{0,0,0}\makebox(0,0)[lt]{\lineheight{1.25}\smash{\begin{tabular}[t]{l}Sim\end{tabular}}}}%
    \put(0,0){\includegraphics[width=\unitlength,page=3]{timesteps_sim_real.pdf}}%
    \put(0.04901655,0.01692713){\color[rgb]{0,0,0}\makebox(0,0)[lt]{\lineheight{1.25}\smash{\begin{tabular}[t]{l}Real\end{tabular}}}}%
    \put(0,0){\includegraphics[width=\unitlength,page=4]{timesteps_sim_real.pdf}}%
    \put(0.13758489,0.00682405){\color[rgb]{0,0,0}\makebox(0,0)[lt]{\lineheight{1.25}\smash{\begin{tabular}[t]{l}t=0.5\end{tabular}}}}%
    \put(0.25334156,0.00682405){\color[rgb]{0,0,0}\makebox(0,0)[lt]{\lineheight{1.25}\smash{\begin{tabular}[t]{l}t=1.5\end{tabular}}}}%
    \put(0.36027149,0.00682405){\color[rgb]{0,0,0}\makebox(0,0)[lt]{\lineheight{1.25}\smash{\begin{tabular}[t]{l}t=2.5\end{tabular}}}}%
    \put(0.46769354,0.00682405){\color[rgb]{0,0,0}\makebox(0,0)[lt]{\lineheight{1.25}\smash{\begin{tabular}[t]{l}t=3.5\end{tabular}}}}%
    \put(0.12388333,0.05320226){\color[rgb]{0,0,0}\makebox(0,0)[lt]{\lineheight{1.25}\smash{\begin{tabular}[t]{l}\textbf{CD=0.032}\\\textbf{HD=0.074}\end{tabular}}}}%
    \put(0.23572481,0.05320226){\color[rgb]{0,0,0}\makebox(0,0)[lt]{\lineheight{1.25}\smash{\begin{tabular}[t]{l}CD=0.088\\HD=0.268\end{tabular}}}}%
    \put(0.34261844,0.05320226){\color[rgb]{0,0,0}\makebox(0,0)[lt]{\lineheight{1.25}\smash{\begin{tabular}[t]{l}\textbf{CD=0.107}\\HD=0.193\end{tabular}}}}%
    \put(0.45028085,0.05320226){\color[rgb]{0,0,0}\makebox(0,0)[lt]{\lineheight{1.25}\smash{\begin{tabular}[t]{l}CD=0.167\\HD=0.389\end{tabular}}}}%
    \put(0.12367481,0.18636942){\color[rgb]{0,0,0}\makebox(0,0)[lt]{\lineheight{1.25}\smash{\begin{tabular}[t]{l}CD=0.045\\HD=0.094\end{tabular}}}}%
    \put(0.23551626,0.18636942){\color[rgb]{0,0,0}\makebox(0,0)[lt]{\lineheight{1.25}\smash{\begin{tabular}[t]{l}CD=0.226\\HD=0.448\end{tabular}}}}%
    \put(0.34240994,0.18636942){\color[rgb]{0,0,0}\makebox(0,0)[lt]{\lineheight{1.25}\smash{\begin{tabular}[t]{l}CD=0.424\\HD=0.646\end{tabular}}}}%
    \put(0.45007235,0.18636942){\color[rgb]{0,0,0}\makebox(0,0)[lt]{\lineheight{1.25}\smash{\begin{tabular}[t]{l}CD=0.281\\HD=0.394\end{tabular}}}}%
    \put(0.12361123,0.32650225){\color[rgb]{0,0,0}\makebox(0,0)[lt]{\lineheight{1.25}\smash{\begin{tabular}[t]{l}CD=0.152\\HD=0.282\end{tabular}}}}%
    \put(0.23545268,0.32510909){\color[rgb]{0,0,0}\makebox(0,0)[lt]{\lineheight{1.25}\smash{\begin{tabular}[t]{l}\textbf{CD=0.069}\\HD=0.262\end{tabular}}}}%
    \put(0.34234635,0.32510909){\color[rgb]{0,0,0}\makebox(0,0)[lt]{\lineheight{1.25}\smash{\begin{tabular}[t]{l}CD=0.185\\HD=0.315\end{tabular}}}}%
    \put(0.45000876,0.32510909){\color[rgb]{0,0,0}\makebox(0,0)[lt]{\lineheight{1.25}\smash{\begin{tabular}[t]{l}CD=0.407\\HD=0.511\end{tabular}}}}%
    \put(0.12374268,0.45827613){\color[rgb]{0,0,0}\makebox(0,0)[lt]{\lineheight{1.25}\smash{\begin{tabular}[t]{l}CD=0.036\\HD=0.108\end{tabular}}}}%
    \put(0.23558413,0.45827613){\color[rgb]{0,0,0}\makebox(0,0)[lt]{\lineheight{1.25}\smash{\begin{tabular}[t]{l}CD=0.088\\\textbf{HD=0.143}\end{tabular}}}}%
    \put(0.34247779,0.45827613){\color[rgb]{0,0,0}\makebox(0,0)[lt]{\lineheight{1.25}\smash{\begin{tabular}[t]{l}CD=0.110\\\textbf{HD=0.146}\end{tabular}}}}%
    \put(0.4501402,0.45827613){\color[rgb]{0,0,0}\makebox(0,0)[lt]{\lineheight{1.25}\smash{\begin{tabular}[t]{l}\textbf{CD=0.136}\\\textbf{HD=0.220}\end{tabular}}}}%
    \put(0.59109096,0.00682405){\color[rgb]{0,0,0}\makebox(0,0)[lt]{\lineheight{1.25}\smash{\begin{tabular}[t]{l}t=1.0\end{tabular}}}}%
    \put(0.70684763,0.00682405){\color[rgb]{0,0,0}\makebox(0,0)[lt]{\lineheight{1.25}\smash{\begin{tabular}[t]{l}t=4.0\end{tabular}}}}%
    \put(0.81377753,0.00682405){\color[rgb]{0,0,0}\makebox(0,0)[lt]{\lineheight{1.25}\smash{\begin{tabular}[t]{l}t=8.0\end{tabular}}}}%
    \put(0.91387459,0.00682405){\color[rgb]{0,0,0}\makebox(0,0)[lt]{\lineheight{1.25}\smash{\begin{tabular}[t]{l}t=14.0\end{tabular}}}}%
    \put(0.5704237,0.05320226){\color[rgb]{0,0,0}\makebox(0,0)[lt]{\lineheight{1.25}\smash{\begin{tabular}[t]{l}CD=0.066\\\textbf{HD=0.130}\end{tabular}}}}%
    \put(0.67390596,0.05320226){\color[rgb]{0,0,0}\makebox(0,0)[lt]{\lineheight{1.25}\smash{\begin{tabular}[t]{l}CD=0.067\\\textbf{HD=0.132}\end{tabular}}}}%
    \put(0.78776541,0.05320226){\color[rgb]{0,0,0}\makebox(0,0)[lt]{\lineheight{1.25}\smash{\begin{tabular}[t]{l}CD=0.062\\HD=0.127\end{tabular}}}}%
    \put(0.90378689,0.05320226){\color[rgb]{0,0,0}\makebox(0,0)[lt]{\lineheight{1.25}\smash{\begin{tabular}[t]{l}CD=0.067\\HD=0.156\end{tabular}}}}%
    \put(0.5702152,0.18636942){\color[rgb]{0,0,0}\makebox(0,0)[lt]{\lineheight{1.25}\smash{\begin{tabular}[t]{l}CD=0.062\\HD=0.142\end{tabular}}}}%
    \put(0.67369739,0.18636942){\color[rgb]{0,0,0}\makebox(0,0)[lt]{\lineheight{1.25}\smash{\begin{tabular}[t]{l}CD=0.061\\HD=0.143\end{tabular}}}}%
    \put(0.78755691,0.18636942){\color[rgb]{0,0,0}\makebox(0,0)[lt]{\lineheight{1.25}\smash{\begin{tabular}[t]{l}\textbf{CD=0.052}\\\textbf{HD=0.122}\end{tabular}}}}%
    \put(0.9035784,0.18636942){\color[rgb]{0,0,0}\makebox(0,0)[lt]{\lineheight{1.25}\smash{\begin{tabular}[t]{l}\textbf{CD=0.041}\\\textbf{HD=0.134}\end{tabular}}}}%
    \put(0.57015161,0.32510909){\color[rgb]{0,0,0}\makebox(0,0)[lt]{\lineheight{1.25}\smash{\begin{tabular}[t]{l}CD=0.058\\HD=0.187\end{tabular}}}}%
    \put(0.67363384,0.32510909){\color[rgb]{0,0,0}\makebox(0,0)[lt]{\lineheight{1.25}\smash{\begin{tabular}[t]{l}CD=0.050\\HD=0.189\end{tabular}}}}%
    \put(0.78749336,0.32510909){\color[rgb]{0,0,0}\makebox(0,0)[lt]{\lineheight{1.25}\smash{\begin{tabular}[t]{l}CD=0.056\\HD=0.144\end{tabular}}}}%
    \put(0.90351485,0.32510909){\color[rgb]{0,0,0}\makebox(0,0)[lt]{\lineheight{1.25}\smash{\begin{tabular}[t]{l}CD=0.170\\HD=0.413\end{tabular}}}}%
    \put(0.57028305,0.45827613){\color[rgb]{0,0,0}\makebox(0,0)[lt]{\lineheight{1.25}\smash{\begin{tabular}[t]{l}\textbf{CD=0.051}\\HD=0.142\end{tabular}}}}%
    \put(0.67376528,0.45827613){\color[rgb]{0,0,0}\makebox(0,0)[lt]{\lineheight{1.25}\smash{\begin{tabular}[t]{l}\textbf{CD=0.049}\\HD=0.151\end{tabular}}}}%
    \put(0.78762481,0.45827613){\color[rgb]{0,0,0}\makebox(0,0)[lt]{\lineheight{1.25}\smash{\begin{tabular}[t]{l}CD=0.082\\HD=0.131\end{tabular}}}}%
    \put(0.90364629,0.45827613){\color[rgb]{0,0,0}\makebox(0,0)[lt]{\lineheight{1.25}\smash{\begin{tabular}[t]{l}CD=0.044\\HD=0.180\end{tabular}}}}%
    \put(0,0){\includegraphics[width=\unitlength,page=5]{timesteps_sim_real.pdf}}%
  \end{picture}%
\endgroup%

%% file: img/time_filter_cdhd_comparison.pdf_tex
\begingroup%
  \makeatletter%
  \providecommand\color[2][]{%
    \errmessage{(Inkscape) Color is used for the text in Inkscape, but the package 'color.sty' is not loaded}%
    \renewcommand\color[2][]{}%
  }%
  \providecommand\transparent[1]{%
    \errmessage{(Inkscape) Transparency is used (non-zero) for the text in Inkscape, but the package 'transparent.sty' is not loaded}%
    \renewcommand\transparent[1]{}%
  }%
  \providecommand\rotatebox[2]{#2}%
  \newcommand*\fsize{\dimexpr\f@size pt\relax}%
  \newcommand*\lineheight[1]{\fontsize{\fsize}{#1\fsize}\selectfont}%
  \ifx\svgwidth\undefined%
    \setlength{\unitlength}{855.23962306bp}%
    \ifx\svgscale\undefined%
      \relax%
    \else%
      \setlength{\unitlength}{\unitlength * \real{\svgscale}}%
    \fi%
  \else%
    \setlength{\unitlength}{\svgwidth}%
  \fi%
  \global\let\svgwidth\undefined%
  \global\let\svgscale\undefined%
  \makeatother%
  \begin{picture}(1,0.87959006)%
    \lineheight{1}%
    \setlength\tabcolsep{0pt}%
    \put(0.14024788,0.0465746){\color[rgb]{0,0,0}\makebox(0,0)[lt]{\lineheight{1.25}\smash{\begin{tabular}[t]{l}10\end{tabular}}}}%
    \put(0.28630137,0.0465746){\color[rgb]{0,0,0}\makebox(0,0)[lt]{\lineheight{1.25}\smash{\begin{tabular}[t]{l}100\end{tabular}}}}%
    \put(0.41461781,0.0465746){\color[rgb]{0,0,0}\makebox(0,0)[lt]{\lineheight{1.25}\smash{\begin{tabular}[t]{l}1000\end{tabular}}}}%
    \put(0.20543175,0.01223021){\color[rgb]{0,0,0}\makebox(0,0)[lt]{\lineheight{1.25}\smash{\begin{tabular}[t]{l}Frequency (Hz)\end{tabular}}}}%
    \put(0.64720875,0.0465746){\color[rgb]{0,0,0}\makebox(0,0)[lt]{\lineheight{1.25}\smash{\begin{tabular}[t]{l}10\end{tabular}}}}%
    \put(0.79326225,0.0465746){\color[rgb]{0,0,0}\makebox(0,0)[lt]{\lineheight{1.25}\smash{\begin{tabular}[t]{l}100\end{tabular}}}}%
    \put(0.92157869,0.0465746){\color[rgb]{0,0,0}\makebox(0,0)[lt]{\lineheight{1.25}\smash{\begin{tabular}[t]{l}1000\end{tabular}}}}%
    \put(0.71239263,0.01223021){\color[rgb]{0,0,0}\makebox(0,0)[lt]{\lineheight{1.25}\smash{\begin{tabular}[t]{l}Frequency (Hz)\end{tabular}}}}%
    \put(0,0){\includegraphics[width=\unitlength,page=1]{time_filter_cdhd_comparison.pdf}}%
    \put(0.04611889,0.52677754){\color[rgb]{0,0,0}\makebox(0,0)[lt]{\lineheight{1.25}\smash{\begin{tabular}[t]{l}$10^{-3}$\end{tabular}}}}%
    \put(0.0455298,0.61364682){\color[rgb]{0,0,0}\makebox(0,0)[lt]{\lineheight{1.25}\smash{\begin{tabular}[t]{l}$10^{-2}$\end{tabular}}}}%
    \put(0.54542651,0.5121716){\color[rgb]{0,0,0}\rotatebox{90}{\makebox(0,0)[lt]{\lineheight{1.25}\smash{\begin{tabular}[t]{l}Simulator Stability $\L_S$\end{tabular}}}}}%
    \put(0.02450186,0.55952679){\color[rgb]{0,0,0}\rotatebox{90}{\makebox(0,0)[lt]{\lineheight{1.25}\smash{\begin{tabular}[t]{l}Step Time (ms.)\end{tabular}}}}}%
    \put(0.02633276,0.24718245){\color[rgb]{0,0,0}\rotatebox{90}{\makebox(0,0)[lt]{\lineheight{1.25}\smash{\begin{tabular}[t]{l}$\text{CD}_d$\end{tabular}}}}}%
    \put(0.54542651,0.24542858){\color[rgb]{0,0,0}\rotatebox{90}{\makebox(0,0)[lt]{\lineheight{1.25}\smash{\begin{tabular}[t]{l}$\text{CD}_q$\end{tabular}}}}}%
    \put(0.04611889,0.70309546){\color[rgb]{0,0,0}\makebox(0,0)[lt]{\lineheight{1.25}\smash{\begin{tabular}[t]{l}$10^{-1}$\end{tabular}}}}%
    \put(0.0455298,0.78377458){\color[rgb]{0,0,0}\makebox(0,0)[lt]{\lineheight{1.25}\smash{\begin{tabular}[t]{l}$10^0$\end{tabular}}}}%
    \put(0.04611889,0.85217647){\color[rgb]{0,0,0}\makebox(0,0)[lt]{\lineheight{1.25}\smash{\begin{tabular}[t]{l}$10^1$\end{tabular}}}}%
    \put(0,0){\includegraphics[width=\unitlength,page=2]{time_filter_cdhd_comparison.pdf}}%
    \put(0.04611889,0.10751204){\color[rgb]{0,0,0}\makebox(0,0)[lt]{\lineheight{1.25}\smash{\begin{tabular}[t]{l}$10^{-1}$\end{tabular}}}}%
    \put(0.05488836,0.19438132){\color[rgb]{0,0,0}\makebox(0,0)[lt]{\lineheight{1.25}\smash{\begin{tabular}[t]{l}$10^0$\end{tabular}}}}%
    \put(0.05488836,0.28382995){\color[rgb]{0,0,0}\makebox(0,0)[lt]{\lineheight{1.25}\smash{\begin{tabular}[t]{l}$10^1$\end{tabular}}}}%
    \put(0.05488836,0.36450907){\color[rgb]{0,0,0}\makebox(0,0)[lt]{\lineheight{1.25}\smash{\begin{tabular}[t]{l}$10^2$\end{tabular}}}}%
    \put(0.05488836,0.43291094){\color[rgb]{0,0,0}\makebox(0,0)[lt]{\lineheight{1.25}\smash{\begin{tabular}[t]{l}$10^3$\end{tabular}}}}%
    \put(0,0){\includegraphics[width=\unitlength,page=3]{time_filter_cdhd_comparison.pdf}}%
    \put(0.56797846,0.52523588){\color[rgb]{0,0,0}\makebox(0,0)[lt]{\lineheight{1.25}\smash{\begin{tabular}[t]{l}$10^{-7}$\end{tabular}}}}%
    \put(0.56738937,0.61210531){\color[rgb]{0,0,0}\makebox(0,0)[lt]{\lineheight{1.25}\smash{\begin{tabular}[t]{l}$10^{-5}$\end{tabular}}}}%
    \put(0.56797846,0.70155397){\color[rgb]{0,0,0}\makebox(0,0)[lt]{\lineheight{1.25}\smash{\begin{tabular}[t]{l}$10^{-3}$\end{tabular}}}}%
    \put(0.56738937,0.79100257){\color[rgb]{0,0,0}\makebox(0,0)[lt]{\lineheight{1.25}\smash{\begin{tabular}[t]{l}$10^{-1}$\end{tabular}}}}%
    \put(0.56797846,0.85063498){\color[rgb]{0,0,0}\makebox(0,0)[lt]{\lineheight{1.25}\smash{\begin{tabular}[t]{l}$10^1$\end{tabular}}}}%
    \put(0,0){\includegraphics[width=\unitlength,page=4]{time_filter_cdhd_comparison.pdf}}%
    \put(-0.00168539,0.47090525){\color[rgb]{0,0,0}\makebox(0,0)[lt]{\lineheight{1.25}\smash{\begin{tabular}[t]{l}a)\end{tabular}}}}%
    \put(0.52301729,0.47089155){\color[rgb]{0,0,0}\makebox(0,0)[lt]{\lineheight{1.25}\smash{\begin{tabular}[t]{l}b)\end{tabular}}}}%
    \put(-0.00073318,0.04610872){\color[rgb]{0,0,0}\makebox(0,0)[lt]{\lineheight{1.25}\smash{\begin{tabular}[t]{l}c)\end{tabular}}}}%
    \put(0.52351737,0.04608133){\color[rgb]{0,0,0}\makebox(0,0)[lt]{\lineheight{1.25}\smash{\begin{tabular}[t]{l}d)\end{tabular}}}}%
    \put(0,0){\includegraphics[width=\unitlength,page=5]{time_filter_cdhd_comparison.pdf}}%
    \put(0.33067736,0.73934528){\color[rgb]{0,0,0}\makebox(0,0)[lt]{\lineheight{1.25}\smash{\begin{tabular}[t]{l}\textbf{\acs{sofa}}\end{tabular}}}}%
    \put(0.33067736,0.77502681){\color[rgb]{0,0,0}\makebox(0,0)[lt]{\lineheight{1.25}\smash{\begin{tabular}[t]{l}\textbf{\acs{flex}}\end{tabular}}}}%
    \put(0.33067736,0.81070836){\color[rgb]{0,0,0}\makebox(0,0)[lt]{\lineheight{1.25}\smash{\begin{tabular}[t]{l}\textbf{\acs{bullet}}\end{tabular}}}}%
    \put(0.33067757,0.84695204){\color[rgb]{0,0,0}\makebox(0,0)[lt]{\lineheight{1.25}\smash{\begin{tabular}[t]{l}\textbf{\acs{mujoco}}\end{tabular}}}}%
    \put(0,0){\includegraphics[width=\unitlength,page=6]{time_filter_cdhd_comparison.pdf}}%
    \put(0.55209678,0.10751204){\color[rgb]{0,0,0}\makebox(0,0)[lt]{\lineheight{1.25}\smash{\begin{tabular}[t]{l}$10^{-1}$\end{tabular}}}}%
    \put(0.56086613,0.19438132){\color[rgb]{0,0,0}\makebox(0,0)[lt]{\lineheight{1.25}\smash{\begin{tabular}[t]{l}$10^0$\end{tabular}}}}%
    \put(0.56086613,0.28382995){\color[rgb]{0,0,0}\makebox(0,0)[lt]{\lineheight{1.25}\smash{\begin{tabular}[t]{l}$10^1$\end{tabular}}}}%
    \put(0.56086613,0.36450907){\color[rgb]{0,0,0}\makebox(0,0)[lt]{\lineheight{1.25}\smash{\begin{tabular}[t]{l}$10^2$\end{tabular}}}}%
    \put(0.56086613,0.43291099){\color[rgb]{0,0,0}\makebox(0,0)[lt]{\lineheight{1.25}\smash{\begin{tabular}[t]{l}$10^3$\end{tabular}}}}%
  \end{picture}%
\endgroup%